\documentstyle[aclap,psfig]{article}

\newenvironment{bib}{\list{}{\setlength{\leftmargin}{.5cm}%
 \setlength{\itemindent}{-.5cm}\setlength{\itemsep}{-1mm}}}{\endlist}

\title{\vspace{-0.5in}Maximizing Top-down Constraints for 
Unification-based Systems}
\author{Noriko Tomuro \\
School of Computer Science, Telecommunications and 
Information Systems \\
DePaul University \\
Chicago, IL 60604 \\
cphdnt@ted.cs.depaul.edu\\ }

\begin{document}

\maketitle
\vspace{-0.5in}
\begin{abstract}

A left-corner parsing algorithm with top-down filtering has been
reported to show very efficient performance for unification-based
systems.  However, due to the nontermination of parsing with 
left-recursive grammars, top-down constraints must be weakened.  
In this paper, a general method of maximizing 
top-down constraints is proposed.
The method provides a procedure to dynamically compute 
$*restrictor*$, a minimum set of features involved in an
infinite loop for every propagation path; thus top-down constraints 
are maximally propagated.
\end{abstract}

\section{Introduction}

A left-corner parsing algorithm with top-down filtering has been
reported to show very efficient performance for unification-based
systems (Carroll, 1994).
In particular, top-down filtering seems to be very effective 
in increasing parse efficiency (Shann, 1991).
Ideally all top-down expectation should be propagated down 
to the input word so that unsuccessful rule applications
are pruned at the earliest time.  
However, in the context of unification-based parsing, left-recursive
grammars have the formal power of a Turing machine, therefore
detection of all infinite loops due to left-recursion is
impossible (Shieber, 1992).  So, top-down constraints must 
be weakened in order for parsing to be guaranteed to terminate.

In order to solve the nontermination problem,
Shieber (1985) proposes {\it restrictor\/}, a statically predefined 
set of features to consider in propagation, and 
{\it restriction\/}, a filtering function which removes 
the features not in {\it restrictor\/} from top-down expectation.
However, not only does this approach fail to provide a
method to automatically generate the restrictor set, 
it may weaken the predicative power of 
top-down expectation more than necessary:
a globally defined {\it restrictor\/} can only specify the least 
common features for all propagation paths.

In this paper, a general method of maximizing 
top-down constraints is proposed.
The method provides a procedure to dynamically compute 
$*restrictor*$, a minimum set of features involved in an
infinite loop, for every propagation path.
Features in this set are selected by the {\it detection function\/},
and will be ignored in top-down propagation.
Using $*restrictor*$, only the relevant features particular
to the propagation path are ignored, thus
top-down constraints are maximally propagated.

\section{Notation}

We use notation from the PATR-II formalism (Shieber, 1986) and
(Shieber, 1992).  Directed acyclic graphs (dags)
are adopted as the representation model.
The symbol $\doteq$ is used to 
represent the equality relation in the unification equations,
and the symbol $\cdot$ used in the form of $p1 \cdot p2$ 
represents the path concatenation of $p1$ and $p2$.

The {\it subsumption\/} relation is defined as ``Dag $D$
{\it subsumes\/} dag $D'$ if $D$ is more general than $D'$.''
The {\it unification\/} of $D$ and $D'$ is notated by $D \sqcup D'$.

The extraction function $D / p1$ extracts the subdag under
path $p1$ for a given $D$, and the embedding function
$D \setminus p1$ injects $D$ into the enclosing dag $D'$ such that
$D' / p1 = D$.  The filtering function $\rho$ is 
similar to (Shieber, 1992):
$\rho(D)$ returns a copy of $D$ in which some features may be
removed.  Note that in this paper $*restrictor*$ specifies
the features to be removed by $\rho$, whereas in 
(Shieber, 1985, 1992) restrictor specifies the features to be 
retained by restriction which is equivalent to $\rho$.

\section{Top-down Propagation}

Top-down propagation can be precomputed to form a 
{\it reachability table\/}.  Each entry in the table is
a compiled dag which represents the relation 
between a non-terminal category and a rule used to
rewrite the constituents in the {\it reachability\/} relation
(i.e., reflexive, transitive closure of the left-corner path).

\begin{figure*}[t]
\centerline{\psfig{figure=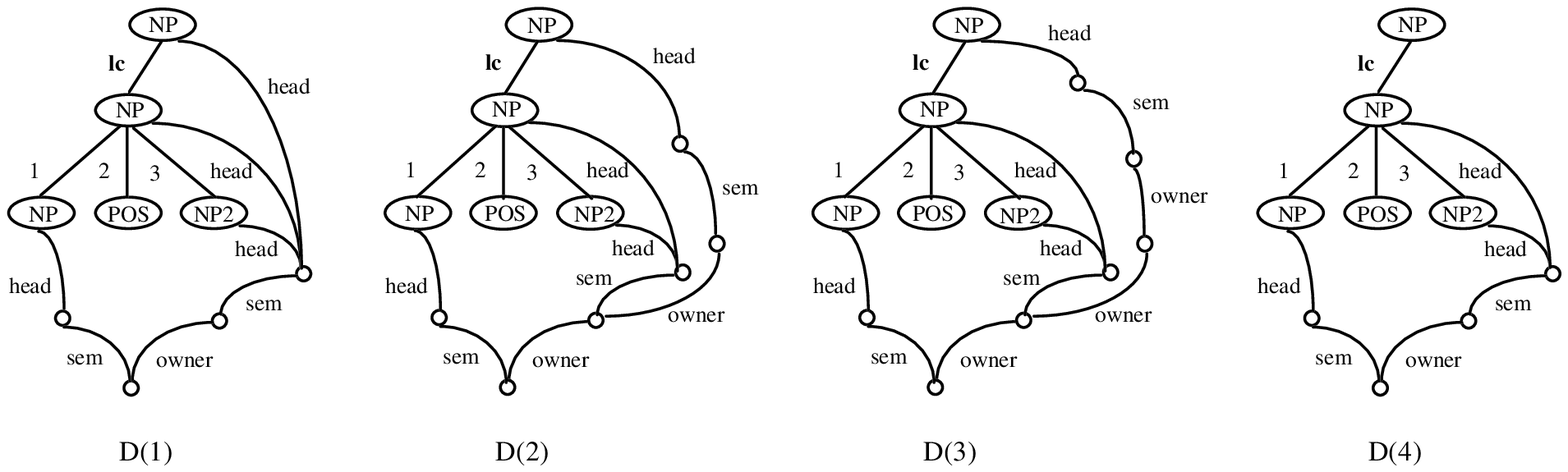,height=2in}}
  \caption{DAGs used in the example}
  \label{mydags}
\end{figure*}

For example, consider the following fragment of a 
grammar used in the syntax/semantics
integrated system called LINK (Lytinen, 1992):

\begin{tabbing}
xxxx\=xxx\=xxx\=xxx\= \kill
$r1:$  \> $NP_{0} \ \rightarrow NP_{1} \ POS \ NP2$ \\
  \> \> $\langle NP_{0} \ head \rangle = \langle NP2 \ head \rangle$ \\
  \> \> $\langle NP_{0} \ head \ sem \ owner \rangle = 
		\langle NP_{1} \ head \ sem \rangle$
\end{tabbing}

(This rule is used to parse phrases such as ``Kris's desk''.)

The dag $D(1)$ in Figure \ref{mydags}\footnote{Category symbols 
are directly indicated in the dag nodes for simplicity.}
represents the initial
application of $r1$ to the category $NP$.  Note that
the subdag under the {\bf lc} arc is the rule used to 
rewrite the constituent on the left-corner path, and
the paths from the top node represent which
top-down constraints are propagated to the lower level.

Top-down propagation works as follows:
given a dag $D$ that represents a reachability relation
and a rule dag $R$ whose left-hand side category (i.e., root)
is the same as $D$'s left-corner category
(i.e., under its ({\bf lc} 1) path), the resulting dag is
$D1 = \rho(D') \sqcup (R \setminus {\bf lc})$, where 
$D'$ is a copy of $D$ in which all the numbered arcs and 
{\bf lc} arc are deleted and the subdag which used to be 
under the ({\bf lc} 1) path is promoted to lie under 
the {\bf lc} arc.  Dags after the next two recursive 
applications of $r1$ ($D(2)$ and $D(3)$ respectively\footnote{In 
this case, $\rho$ is assumed to be 
an identity function.}) are shown in Figure \ref{mydags}.

Notice the filtering function $\rho$ is applied only 
to $D'$.  In the case when $\rho(D') = nil$, the top node in 
$D1$ will have no connections to the
rule dag under the {\bf lc} arc.  This means no top-down constraints
are propagated to the lower level, therefore the parsing
becomes pure bottom-up.

In many unification-based systems, subsumption is used
to avoid redundancy: a dag is recorded in the table if it is not 
subsumed by any other one.  Therefore, if a newly created dag
is incompatible or more general than existing dags,
rule application continues.  In the above example,
$D(2)$ is incompatible with $D(1)$ and therefore gets entered
into the table.  The $owner$ arc keeps extending
in the subsequent recursive applications (as in $D(3)$), thus the 
propagation goes into an infinite loop.  

\subsection{Proposed Method}

Let $A$ be a dag created by the first application of the rule
$R$ and $B$ be a dag created by the second application during
the top-down propagation.\footnote{In the case of indirect recursion,
there are some intervening rule applications between $A$ and $B$.}
In the proposed method,
$A$ and $B$ are first checked for subsumption.
If $B$ is subsumed by $A$, the propagation for this path
terminates.  Otherwise a possible loop is detected.
The detection function (described in the next subsection) 
is called on $A$ and $B$
and selected features are added to the $*restrictor*$ 
set.\footnote{A separate $*restrictor*$ must be kept for each
propagation path.}
Then, using the updated $*restrictor*$, propagation is 
re-done from $A$.

When $R$ is applied again yielding $B'$,
while $B'$ is not subsumed by $A$,
the following process is repeated:
if $B'$ is incompatible with $A$, the detection function is
called on $A$ and $B'$ and propagation is re-done from $A$.
If $B'$ is more general than $A$, then $A$ is replaced by $B'$
(thereby keeping the most general dag for the path)
and propagation is re-done from $B'$.
Otherwise the process stops for this propagation path.
Thus, the propagation will terminate 
when enough features are
detected, or when $*restrictor*$ includes all the (finite 
number of) features in the grammar.\footnote{In reality, category 
feature will never be in $*restrictor*$ because the same rule $R$ 
is applied to derive both $A$ and $B'$.}

In the example, when the detection function
is called on $D(1)$ and $D(2)$ after the first recursive
application, the feature {\it owner\/} is selected 
and added to $*restrictor*$.  After the 
propagation is re-done from $D(1)$, the resulting dag $D(4)$
becomes more general than $D(1)$.\footnote{Remember 
$D(4) = \rho(D(1)') \sqcup (r1 \setminus {\bf lc})$
where $\rho$ filters out {\it owner\/} arc.}
Then $D(1)$ is replaced
by $D(4)$, and the propagation is re-done once again.
This time it results the same $D(4)$, therefore the
propagation terminates.

\subsection{Detection Function}

The detection function compares two dags $X$ and $Y$ by
checking every constraint (unification equation) $x$ in $X$ 
with any inconsistent or more general constraint $y$ in $Y$.
If such a constraint is found, the function selects a path
in $x$ or $y$ and detects its last arc/feature as being 
involved in the possible loop.\footnote{This scheme may be 
rather conservative.}

If $x$ is the path constraint $p1 \doteq p2$ where 
$p1$ and $p2$ are paths of length $\geq 1$, features may be
detected in the following cases:\footnote{Note the cases in 
this section do {\bf not} represent all possible situations.}
\begin{itemize}
  \item (case 1) If both $p1$ and $p2$ exist in $Y$, and there 
	exists a more general constraint $y$ in $Y$ in the form 
	$p1 \cdot p3 \doteq p2 \cdot p3$ (length of $p3$ is 
	also $\geq 1$), the path $p3$ is selected;
  \item (case 2) If both $p1$ and $p2$ exist in $Y$, but the 
	subdag under $p1$ and the subdag under $p2$ do not unify, 
	or if neither $p1$ nor $p2$ exists in $Y$, 
	whichever of $p1$ or $p2$ does not contain the {\bf lc} 
	arc, or either if they both contain the {\bf lc} arc, 
	is selected; and
  \item (case 3) If either $p1$ or $p2$ does not exist in $Y$, 
	the one which does not exist in $Y$ is selected.
\end{itemize}
If $x$ is the constant constraint $p1 \doteq c$ (where $c$
is some constant),
features may be detected in the following cases:
\begin{itemize}
  \item (case 4) If there exists an incompatible constraint $y$ 
	of the form $p1 \doteq d$ where $d \neq c$ in $Y$,
	or if there is no path $p1$ in $Y$, 
	$p1$ is selected; and
  \item (case 5) If there exists an incompatible constraint $y$
	of the form $p1 \cdot p2 \doteq c$, then $p2$ is 
	selected.
\end{itemize}

\section{Related Work}

A similar solution to the nontermination problem with unification 
grammars in Prolog is proposed in (Samuelsson, 1993).  
In this method, an operation called anti-unification (often 
referred to as {\it generalization\/} as the counterpart of 
unification) is applied to
the root and leaf terms of a cyclic propagation,
and the resulting term is stored in the reachablity table
as the result of applying restriction on both terms.
Another approach taken in (Haas, 1989) eliminates the
cyclic propagation by replacing the features 
in the root and leaf terms with new variables.

The method proposed in this paper is more
general than the above approaches: if the selection ordering
is imposed in the detection function, 
features in $*restrictor*$ can be 
collected incrementally as the cyclic propagations are repeated.
Thus, this method is able to create a less restrictive $*restrictor*$
than these other approaches.

\section{Discussion and Future Work}

The proposed method has an obvious difficulty:
the complexity caused by the repeated propagations could 
become overwhelming for some grammars.  
However, in the experiment on LINK system using a fairly broad 
grammar (over 130 rules), precompilation terminated
with only a marginally longer processing time.

In the experiment, all features (around 40 syntactic/semantic 
features) except for one in the example in this paper were
able to be used in propagation.  In the preliminary analysis,
the number of edges entered into the chart has decreased by 30\%
compared to when only the category feature (i.e., context-free
backbone) was used in propagation.

For future work, we intend to apply the proposed method to
other grammars.  By doing the empirical analysis of precompilation 
and parse efficiency for different grammars, we will be able to 
conclude the practical applicability of the proposed method.
We also indend to do more exhaustive case analysis
and investigate the selection ordering of 
the detection function.  Although the current 
definition covers most cases, it is by no means complete.

\vspace{.4cm}
\noindent 
\large {\bf References}
\normalsize
\begin{bib}
\item Carroll, J. \ (1994). Relating complexity to practical
	performance in parsing with wide-coverage unification 
	grammars. In {\it Proceedings of the 32nd Annual 
	Meeting of the Association for Computational Linguistics},
	pp.\ 287-293.

\item Haas, A. \ (1989). 
	A parsing algorithm for unification grammar,
	{\it Computational Linguistics\/}, {\bf 15}(4), 
	pp.\ 219-232.

\item Lytinen, S.\ (1992).  A unification-based, integrated 
	natural language processing system.
	{\em Computers and Mathematics with Applications\/},
	{\bf 23}(6-9), pp.\ 403-418.


\item Samuelsson, C. \ (1993). Avoiding non-termination in
	unification grammars.  In {\it Proceedings of Natural 
	Language Understanding and Logic Programming IV}, 
	Nara, Japan.

\item Shann, P.\  (1991).  Experiments with GLR and chart parsing.
	In Tomita, M. {\em Generalized LR Parsing.}  Boston:  
	Kluwer Academic Publishers, p.\ 17-34.

\item Shieber, S. \ (1985). Using restriction to extend parsing
	algorithms for complex-feature-based formalisms.
	In {\it Proceedings of the 23rd Annual Meeting of the 
	Association for Computational Linguistics}, Chicago, IL,
	pp.\ 145-152.

\item Shieber, S.\  (1986). {\em  An Introduction to 
	Unification-Based Approaches to Grammar.} Stanford, CA: 
	Center for the Study of Language and Information.

\item Shieber, S.\ (1992). {\it Constraint-based Grammar 
	Formalisms\/}. Cambridge, MA: MIT Press.
\end{bib}
\end{document}